\documentclass[10pt,twocolumn,letterpaper,table]{article}

\usepackage{iccv}
\usepackage{times}
\usepackage{epsfig}
\usepackage{graphicx}
\usepackage{amsmath,bm}
\usepackage{amssymb}
\usepackage{enumitem}
\usepackage{xcolor}
\usepackage{color}
\usepackage{multirow}
\usepackage{authblk}
\usepackage[pagebackref=true,breaklinks=true,letterpaper=true,colorlinks,bookmarks=false]{hyperref}

\iccvfinalcopy % *** Uncomment this line for the final submission

\def\iccvPaperID{2249} % *** Enter the CVPR Paper ID here

\ificcvfinal\fi

\makeatletter
\renewcommand\AB@affilsepx{  \protect\Affilfont}
\makeatother
\begin{document}
	
	%%%%%%%%% TITLE
	\title{Rotation equivariant vector field networks}
    
    \author[1]{Diego Marcos \thanks{Corresponding author: diego.marcos@geo.uzh.ch}}
    \author[1]{Michele Volpi}
    \author[2]{Nikos Komodakis}
    \author[1]{Devis Tuia}
    \affil[1]{University of Zurich, }\affil[2]{Ecole des Ponts, Paris Tech \authorcr}
    
    %\affil[ ]{{\tt\small diego.marcos@geo.uzh.ch michele.volpi@geo.uzh.ch nikos.komodakis@enpc.fr devis.tuia@geo.uzh.ch}}

% 	\author{Diego Marcos\\
% 		University of Zurich\\
% 		%Switzerland\\
% 		{\tt\small diego.marcos@geo.uzh.ch}
% 		% For a paper whose authors are all at the same institution,
% 		% omit the following lines up until the closing ``}''.
% 		% Additional authors and addresses can be added with ``\and'',
% 		% just like the second author.
% 		% To save space, use either the email address or home page, not both
% 		\and
% 		Michele Volpi\\
% 		University of Zurich\\
% 		%Switzerland\\
% 		{\tt\small michele.volpi@geo.uzh.ch}
% 		\and
% 		Nikos Komodakis\\
% 		Ecole des Ponts, Paris Tech\\
% 		%France\\
% 		{\tt\small nikos.komodakis@enpc.fr}
% 		\and
% 		Devis Tuia\\
% 		University of Zurich\\
% 		%Switzerland\\
% 		{\tt\small devis.tuia@geo.uzh.ch}
% 	}
	
\maketitle
%\thispagestyle{empty}
	
%%%%%%%%% ABSTRACT
\begin{abstract}
In many computer vision tasks, we expect a particular behavior of the output with respect to rotations of the input image. If this relationship is explicitly encoded, instead of treated as any other variation, the complexity of the problem is decreased, leading to a reduction in the size of the required model. 
  
In this paper, we propose the \emph{Rotation Equivariant Vector Field Networks (RotEqNet)}, a Convolutional Neural Network (CNN) architecture encoding rotation equivariance, invariance and covariance.
Each convolutional filter is applied at multiple orientations and returns a vector field representing magnitude and angle of the highest scoring orientation at every spatial location. %A modified convolution operator using vector fields as inputs and filters can then be applied to obtain deep architectures. 
We develop a modified convolution operator relying on this representation to obtain deep architectures. 
We test RotEqNet on several problems requiring different responses with respect to the inputs' rotation: image classification, biomedical image segmentation, orientation estimation and patch matching. In all cases, we show that RotEqNet offers extremely compact models in terms of number of parameters and provides results in line to those of networks orders of magnitude larger.
\end{abstract}
\vspace*{-8mm}
	
\section{Introduction}
	
In many real life problems, such as overhead (aerial or satellite) or biomedical image analysis, there are no dominant up-down or left-right relationships. For example, when detecting cars in aerial images, the object's absolute orientation is not a discriminant feature. %Only the \emph{relative} orientation to surrounding elements such as roads and buildings could provide additional prior information.
If the absolute orientation of the image is changed, e.g. by following a different flightpath, we would expect the car detector to score the exact same values over the same cars, just in their new position on the rotated image, independently from their new orientation along the image axes. In this case, we say that the problem is rotation \emph{equivariant}: rotating the input is expected to result in the same rotation in the output. On the other hand, if we were confronted with a classification setting in which we are only interested in the presence or absence of cars in the whole scene, the classification score should remain the same, no matter the absolute orientation of the input scene. In this case the problem is rotation \emph{invariant}. The more general case would be rotation \emph{covariance}, in which the output changes as a function of the rotation of the input, with some predefined behavior. Taking again the cars example, a rotation covariant problem would be to retrieve the absolute orientation of cars with respect to longitude and latitude: in this case, a rotation of the image should produce a change of the predicted angle. % of the cars: by rotating the input image the output would change as a function of the rotation angle.
	
\begin{figure}[!t]
\centering
\includegraphics[width=0.95\linewidth]{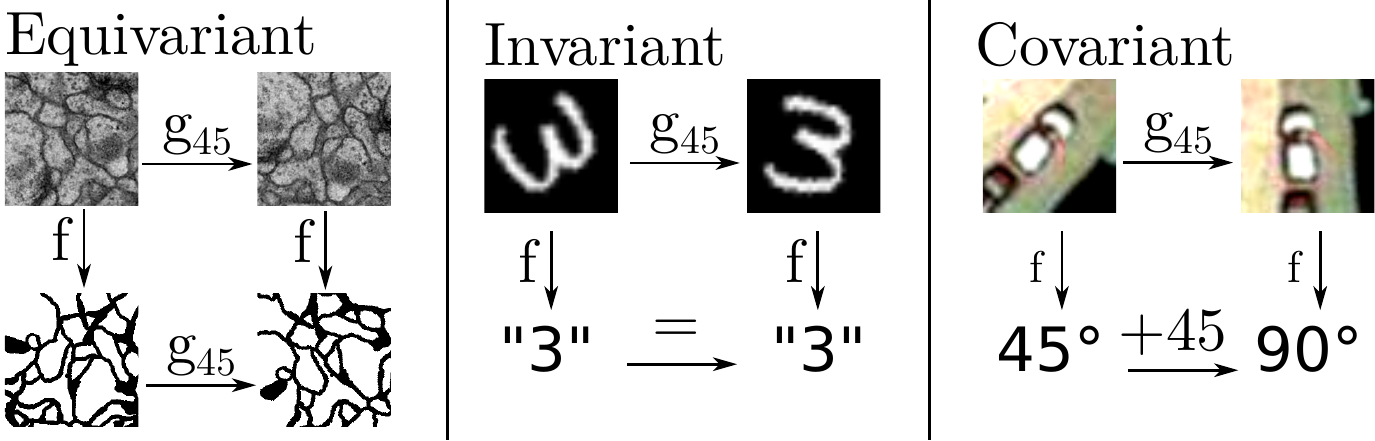}
\vspace*{-0mm}
\caption{Desirable behaviors with respect to rotation of the inputs: (left) equivariance in segmentation; (center) invariance in classification; (right) covariance in absolute orientation estimation. g$_{45}$ is an operator that rotates the input image by $45^\circ$. }
\label{fig:equi}
\vspace*{-4mm}
\end{figure}
	
Throughout this article we will make use of the terms equivariance, invariance and covariance of a function $f(\cdot)$ with respect to a transformation $g(\cdot)$ in the following sense:
%\vspace*{-2mm}
\begin{description}[noitemsep]
\setlength\itemsep{0.2mm}
\item[\ \  - equivariance:] $f\large(g(\cdot)\large) = g\large(f(\cdot)\large)$,
\item[\ \  - invariance:] $f\large(g(\cdot)\large) = f(\cdot)$,
\item[\ \  - covariance:] $f\large(g(\cdot)\large) = g'\large(f(\cdot)\large)$,
\end{description}
%\vspace*{-2mm}    
where $g'(\cdot)$ is a second transformation, which is itself a function of $g(\cdot)$. With the above definitions, equivariance and invariance are special cases of covariance. We illustrate these properties in Fig.~\ref{fig:equi}.

%Although these 3 properties have often been heavily exploited in many problems, it is not so for the rotation transformation, which has only recently started to gather more interest.
	
In this paper, we propose a CNN architecture that naturally encodes these three properties: RotEqNet.
In the following, we will recall how CNNs achieve translation invariance, before discussing our own proposition.

\begin{figure*}[!t]
\centering
\includegraphics[width=0.95\linewidth]{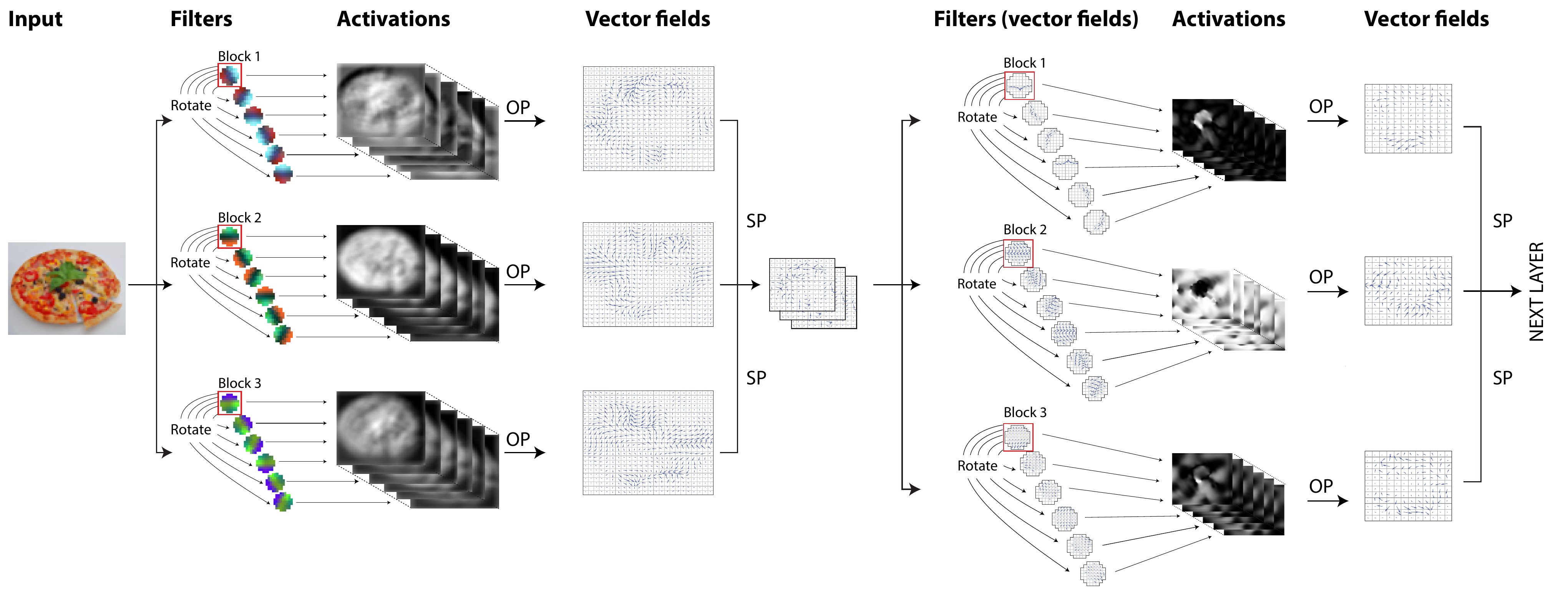}
\vspace*{-2mm}
\caption{Example of the first two layers of RotEqNet. Each layer learns only three canonical filters (red squares) and replicates them across six orientations. The output of the first block are three vector field maps, which are further convolved by vector field filters in the second block (OP: orientation pooling; SP: spatial pooling).}
\label{fig:layer1}
\vspace*{-4mm}
\end{figure*}

\subsection{Dealing with translations in CNNs}

The success of CNNs is partly due to the translation equivariant nature of the convolution operation. 
    %To understand how to achieve equivariance to rotations by means of convolutions, we will briefly summarize how translation equivariance is obtained by standard CNNs. \devis{}{[ALREADY SAID. WE CAN REMOVE IF WE NEED SPACE.]}
%
%\vspace*{-0.3 cm}
%
%\paragraph{Translation equivariance in CNNs:}
The convolution of an image $\mathbf{x}\in \mathbb{R}^{M\times N\times d}$ with a filter $\textbf{w}\in \mathbb{R}^{m\times n\times d}$, written $\mathbf{y}=\mathbf{w}*\mathbf{x}$, is obtained by applying the same scalar product operation over all overlapping $m\times n$ windows (unit stride) on $\mathbf{x}$. If $\mathbf{x}$ undergoes an integer translation in the horizontal and vertical directions by $(p,q)$ pixels, the same pixel neighborhoods in $\mathbf{x}$ will exist in the translated $\mathbf{x}$, but again translated by $(p,q)$ pixels. Therefore, any operation involving fixed neighborhoods
such as the convolution is translation equivariant.% when applied to images.%, since these are characterized by local neighborhoods with identical arrangement.

A crucial consequence of learning convolution weights is a drastic reduction in the number of parameters. Without the translation equivariance assumption, each local window would have a different set of weights.
Forcing weights to be shared across locations, known as \emph{weight tying}, reduces the number of learnable parameters proportionally to the number of pixels in the image and hardcodes translation equivariance within the model. This fact is vital for the applicability of deep neural networks to images~\cite{le1990handwritten}.

\subsection{Incorporating rotation equivariance in CNNs} \label{sec:expDim}
    
RotEqNet shows similar advantageous characteristics when dealing with rotations: by encoding equivariance, we are able to strongly reduce the number of parameters while keeping similar or better accuracy across different tasks.
	
However, applying the exact same reasoning of weight tying for rotations is not straightforward. To follow the same logic, one should apply $R$ rotated versions of each convolutional filter, resulting in $R$ feature maps per filter. The dimensionality of subsequent filters would therefore increase with $R$, strongly increasing model size and requirements for runtime memory usage.
	
One way of reducing the size of the model while keeping rotation equivariance would be to propagate only the maximum value occurring across $R$ feature maps. However, deeper layers would have no information about the orientation of features at previous layers.

We propose a trade-off between these two approaches by keeping the maximum value across the $R$ feature maps, but in the form of a 2D vector field that captures its \emph{magnitude} and \emph{orientation} and propagates it through all the layers of the network.%: we call this approach Rotation Equivariant Vector Field Network (RotEqNet).

\section{Related work}

Two families of approaches explicitly account for rotation invariance or equivariance: 1) those that transform the representation (image or feature maps) and 2) those that rotate the filters. RotEqNet belongs to the latter.
	
\vspace*{-3mm}
	
\paragraph{1) Rotating the inputs:}
Jaderberg~\etal~\cite{jaderberg2015spatial} propose the Spatial Transformer layer, which learns how to crop and transform a region of the image (or a feature map) before passing it to the next layer. This transforms relevant regions into a canonical form, improving the learning process by reducing geometrical appearance variations in subsequent layers.
TI-pooling \cite{laptev2016ti} inputs several rotated versions of a same image to the same CNN and then performs pooling across the different feature vectors at the first fully connected layer. Such scheme allows another subsequent fully connected layer to choose among rotated inputs to perform classification. 
Cheng~\etal~\cite{cheng2016rifd} employ in every minibatch several rotated versions of the input images. Their representations after the first fully connected layer are then encouraged to be similar, forcing the CNN to learn rotation invariance.
Henriques~\etal~\cite{henriques2016warped} warp the images such that the translation equivariance inherent to convolutions is transformed into rotation and scale equivariance.

On the one hand, these methods have the advantage of exploiting conventional CNN implementations, since they only act on data representations. On the other hand, they can only consider global transformations of the input images. While this is well suited for tasks such as image classification, it limits their applicability to other problems (e.g. semantic segmentation), where the local relative orientation of certain objects with respect to surroundings is what matters. Instead, RotEqNet is based on specific CNN building blocks designed to deal with local orientation information. Therefore, RotEqNet can approach diverse tasks such as classification, fully convolutional semantic segmentation, detection and regression.

It is worth mentioning that standard data augmentation strategies belong to this first family. They rely on random rotations and flips of the training samples~\cite{simard2003best}: given abundant training samples and enough model capacity, a CNN might learn that different orientations should score the same by learning equivalent filters at different orientations~\cite{lenc2015understanding}. Unlike this, RotEqNet is well suited for problems with limited training samples that can profit from reduced model sizes, since the behavior with respect to rotations is hardcoded and it does not need to be learned.

\vspace*{-3mm}
\paragraph{2) Rotating the filters:}
Gens and Domingos~\cite{gens2014deep} tackle the problem of the exploding dimensionality (discussed in Sec.~\ref{sec:expDim}) by applying learnable pooling operations and sampling the symmetry space at each layer. This way, they avoid applying the filters exhaustively across the (high dimensional) feature maps by selectively sampling few rotations. By doing so, only the least important information is lost from layer to layer.  Cohen~\etal~\cite{cohen2016group,cohen2016steerable} use a smaller symmetry group, composed of a flipping and four $90^\circ$ rotations and perform pooling within the group. They apply it only in deeper layers, since they found that pooling in the early layers discards important information and harms the performance. 
Instead of explicitly defining a symmetry group, Ngiam~\etal~\cite{ngiam2010tiled} pool across several untied filters, thus letting the network learn the type of invariance. Sifre~\etal~\cite{sifre2013rotation} use hand crafted wavelets that are separable in the roto-translational space, allowing for more efficient computations. 
Another approach to avoid the dimensionality explosion is to limit the depth of the network: Sohn~\etal~\cite{sohn2012learning} and Kivinen~\etal~\cite{kivinen2011transformation} propose such a scheme with Restricted Boltzmann Machines (RBM), while Marcos~\etal~\cite{marcos2016learning} consider supervised CNNs consisting of a single convolutional layer.
	
These works find a compromise between the computational resources required and the amount of orientation information kept throughout the layers, by either keeping the model shallow or accounting for a limited amount of orientations. With RotEqNet, we avoid such compromise by pooling multiple orientations and passing forward both the maximum magnitude \emph{and} the orientation at which it occurred. This modification allows to build deep rotation equivariant architectures, in which deeper layers are aware of the dominant orientations. At the same time, the dimensionality of feature maps and filters is kept low by discarding information about non-maximum orientations, thus reducing memory requirements.
	
The most similar approaches to RotEqNet are the recently proposed Harmonic Networks (H-Nets)~\cite{worrall2016harmonic} and Oriented Response Networks (ORN)~\cite{zhou2017oriented}, both of which use an enriched feature map explicitly capturing the underlying orientations. They do so by using either complex circular harmonics (H-Nets) or the full vector of oriented responses (ORN). H-Nets offer a very compact feature map, but are limited to learning filters that are a combination of circular harmonic wavelets. On the other hand, ORN allows to learn arbitrary filters, but relies on a much less compact representation of the feature maps, leading to heavier models both in terms of size and memory requirements. RotEqNet provides the best of both worlds: the compactness of the former with the flexibility of the latter. These properties make it particularly suitable to address problems characterized by limited training samples, as we will see in the experiments.

\section{Rotation equivariant vector field networks}
	
We focus on achieving rotation equivariance by performing convolutions with several rotated instances of the same \emph{canonical} filter (see Fig.~\ref{fig:layer1}). The canonical filter $\mathbf{w}$ is rotated at $R$ different evenly spaced orientations.%, evenly spaced in a given interval of angles.
In the experiments (Sec.~\ref{sec:exp}) we deal with problems requiring either full invariance, equivariance or covariance, so we use the interval $\boldsymbol{\alpha} = [0^\circ, 360^\circ]$. However, this interval can be adapted to a  known range of tilts.
The output of the filter $\mathbf{w}$ at a specific location consists of the magnitude of the maximal activation across the orientations and the corresponding angle. If we convert this polar representation into Cartesian coordinates, each filter $\mathbf{w}$ produces a vector field feature map $\mathbf{z}\in \mathbb{R}^{H\times W \times 2}$, where the output of each location consists of two values $[u,v]\in \mathbb{R}^2$ implicitly encoding the maximal activation in both magnitude and direction. Since the feature maps have become vector fields, from this moment on the filters must also be vector fields, as seen in the right part of Fig.~\ref{fig:layer1}.
	
The advantage of representing $\mathbf{z}$ in Cartesian coordinates is that the horizontal and vertical components $[u,v]$ are orthogonal, and thus a convolution of the two vector fields can be computed on each component independently using standard convolutions (see Eq. (\ref{eq:orth})).

\subsection{RotEqNet building blocks}
	
RotEqNet requires specific building blocks to handle vectors fields as inputs and/or outputs (Fig.~\ref{fig:layer1}). In the following, we present our reformulation of traditional CNN blocks to account for both vector field activations and filters. The implementation\footnote{Will be made available at http://github.com/di-marcos/RotEqNet} is based on the MatConvNet~\cite{vedaldi15matconvnet} toolbox\footnote{http://www.vlfeat.org/matconvnet}.
	
\subsubsection{Rotating convolution (RotConv)}\label{met:rot_conv}
	
Given an input image with $m/2$ zero-padding $\mathbf{x}\in \mathbb{R}^{H+m/2\times W+m/2 \times d}$, we apply the filter $\mathbf{w}\in \mathbb{R}^{m\times m \times d}$ at $R$ orientations, corresponding to the angles:
\begin{equation}
	\alpha_r = \frac{360}{R}r \quad \forall r = 1,2\dots R. 
\end{equation}
Each one of these rotated versions of the canonical filters (highlighted by red squares in Fig.~\ref{fig:layer1}) is computed by resampling $\mathbf{w}$ with bilinear interpolation after rotation of $\alpha_r$ degrees around the filter's center.
\begin{equation}
	\mathbf{w}^{r} = g_{\alpha_r}(\mathbf{w}),
\end{equation}
where $g_\alpha$ is the $\alpha$ degrees rotation operator. Interpolation is always required unless only rotations of multiples of $90^\circ$ are considered. In practice, this means that the rotation equivariance will only be approximate. 
	
%The position $[i',j']$ after rotation of a specific filter weight, originally located at $[i,j]$ in the canonical form, is% computed as:
	%
%\begin{equation}
%	[i',j'] = [i,j]
%	\bigg[
%	\begin{array}{c c}
%		\cos(\alpha_r) & \sin(\alpha_r)\\
%		-\sin(\alpha_r) & \cos(\alpha_r)
%	\end{array}
%	\bigg].
%\end{equation}
%
%Coordinates are relative to the center of the filter. 
Since the rotation can force weights near the corners of the filter to be relocated outside of its spatial support, only the weights within a circle of diameter $m$ pixels are used to compute the convolutions. The output tensor $\mathbf{y}\in \mathbb{R}^{H\times W\times R}$  consists of $R$ feature maps computed as:
\begin{equation}
\mathbf{y}^{(r)} = (\mathbf{x}\ast\mathbf{w}^{r}) \quad \forall r = 1,2\dots R,
\end{equation}
where $(\ast)$ is the convolution operator.
%such that
%
%\begin{equation}
%(\mathbf{x}\ast\mathbf{w})[i,j] = \sum_m\sum_n \mathbf{x}[i-m,j-n]\mathbf{w}[m,n],
%\end{equation}
%
%where $[m,n]$ is the neighborhood considered by the filter. }{.}
The tensor $\mathbf{y}$ encodes the roto-translation output space such that rotation in the input corresponds to a translation across the feature maps. Note that only the canonical filter $\mathbf{w}$ is actually stored in the model. During backpropagation, gradients corresponding to each rotated filter $\nabla\mathbf{w}^{r}$ are aligned back to the canonical form and added: 
\begin{equation}
\nabla \mathbf{w} = \sum_r g_{-\alpha_r}(\nabla\mathbf{w}^{r}).
\end{equation}
This block can be applied on conventional CNN feature maps (left side of Fig.~\ref{fig:layer1}) or on vector field feature maps (right side of Fig.~\ref{fig:layer1}). In the second case it is computed on each component \emph{independently} and the resulting 3D tensors added:
\begin{equation}
(\mathbf{z}\ast\mathbf{w}) = (\mathbf{z}_u\ast\mathbf{w}_u) +(\mathbf{z}_v\ast\mathbf{w}_v),\label{eq:orth}
\end{equation}
where subscripts $u$ and $v$ denote the horizontal and vertical components. 
	
It is important to note that the image rotation operator $g_{\alpha}$ requires an additional step when $\mathbf{w}\in \mathbb{R}^{m\times m \times 2}$ is a 2D vector field. The components of $\mathbf{w}^r=g_{\alpha_r}(\mathbf{w})$ have to be computed as:
\begin{align}
\mathbf{w}^r_u = cos(\alpha_r)g_{\alpha_r}(\mathbf{w}_u) - sin(\alpha_r)g_{\alpha_r}(\mathbf{w}_v)\\ 
\mathbf{w}^r_v = cos(\alpha_r)g_{\alpha_r}(\mathbf{w}_v) + sin(\alpha_r)g_{\alpha_r}(\mathbf{w}_u)
\end{align}
\subsubsection{Orientation pooling (OP):}

Given the output 3D tensor $\mathbf{y}$, the role of the orientation pooling is to convert it to a 2D vector field $\mathbf{z} \in \mathbb{R}^{H\times W\times 2}$. This avoids the exploding dimensionality problem by only keeping information about the maximally activating orientation of $\mathbf{w}$. First, we extract a 2D map of the largest activation magnitudes, $\bm{\rho} \in \mathbb{R}^{H\times W}$, and their corresponding orientations, $\bm{\theta} \in \mathbb{R}^{H\times W}$. Specifically, for activations located at $[i,j]$:
\begin{align}
\bm{\rho}[i,j] = & \max_{r} \mathbf{y}[i,j,r], \\
\bm{\theta}[i,j] = &\frac{360}{R}\, \displaystyle\operatorname{arg\,max}_{r} \mathbf{y}[i,j,r].
\end{align}
This can be treated as a polar representation of a 2D vector field as long as $\bm{\rho}[i,j] \ge 0 \quad \forall i,j$, a condition that is met when using any function on $\mathbf{y}$ that returns non-negative values prior to the OP. We employ the common Rectified Linear Unit (ReLu) operation, defined as $\texttt{ReLu}(x)=\max(x,0)$, to $\bm{\rho}$, as it provides non-saturating, sparse nonlinear activations offering stable training. Then, this representation can be transformed into Cartesian coordinates as:
\begin{align}
	\mathbf{u} = \texttt{ReLu}(\bm{\rho}) \cos(\bm{\theta}) \\ 
	\mathbf{v} = \texttt{ReLu}(\bm{\rho}) \sin(\bm{\theta})
\end{align}
with $\mathbf{u},\mathbf{v}\in\mathbb{R}^{H\times W}$. The 2D vector field $\mathbf{z}$ is then built as:
\begin{equation}
\mathbf{z} = \bigg[\begin{tabular}{c}
	1 \\ 
	0 \\ 
\end{tabular}  \bigg] \mathbf{u} + 
\bigg[\begin{tabular}{c}
	0 \\ 
	1 \\ 
\end{tabular}  \bigg] \mathbf{v}
\end{equation}
	
%\paragraph{Orientation un-pooling:}
	
%The adjoint of the orientation pooling is the orientation un-pooling. It takes a vector field $\mathbf{z}$ and outputs a stack of scalar maps $\mathbf{y}$, where each individual map $\mathbf{y}^{(r)}$ is such that:
	%
	%\begin{equation}
	%	\mathbf{y}^{(r)} = \bm{\rho} \cdot \bigg[\alpha_r - \frac{360}{2R} < \bm{\theta} \ge \alpha_r + \frac{360}{2R}\bigg], r\in 1\ldots R.
	%\end{equation}
	%
	%The resulting $\mathbf{y}$ will contain mostly zeros and its values will be such that the result of applying an orientation pooling operator to $\mathbf{y}$ will result in $\mathbf{z}$. The orientation un-pooling block is required for the backpropagation of gradients through an OP block. It can also be used for rotation equivariant up-sampling after a RotConv$^\top$ layer, to convert vector field $\mathbf{z}$ into roto-translation $\mathbf{y}$ feature maps. An alternative up-sampling is to use a RotConv operator directly on $\mathbf{z}$.
	
\subsubsection{Spatial pooling (SP) for vector fields}
	
Max-pooling is commonly used in CNNs to obtain some invariance to small deformations and reducing the size of the feature maps. This is done by downsampling the input feature map $\mathbf{x}\in \mathbb{R}^{M\times N\times d}$ to $\mathbf{x}_p\in \mathbb{R}^{\frac{M}{p}\times \frac{N}{p}\times d}$. This operation is performed by taking the maximum value contained in each one of the $C$ non-overlapping $p\times p$ regions of $\mathbf{x}$, indexed by $c$. It is computed as $\mathbf{x}_p[c] = \max_{i \in c} \mathbf{x}[i]$, which can be expressed as:
\begin{equation}
\mathbf{y}_p[c] =\mathbf{y}[j], \text{ where } j = \operatorname{arg\,max}_{i \in c}\mathbf{y}[i].
\end{equation}
This allows us to define a max-pooling for vector fields as:%
\begin{equation}
\mathbf{z}_p[c]=\mathbf{z}[j], \text{ where } j = \displaystyle\operatorname{arg\,max}_{i \in c}\boldsymbol{\rho}[i],
\end{equation}
where $\bm{\rho}$ is a standard scalar map containing the magnitudes of the vectors in $\mathbf{z}$.
	
\subsubsection{Batch normalization (BN) for vector fields}
		
BN~\cite{ioffe2015batch} normalizes every feature map in a mini-batch to zero mean and unit standard deviation. It improves convergence by training with stochastic gradient descent.
	
In our case, since working with vector fields of magnitude and orientation of activations, %the feature maps are not scalar fields, but vector fields representing the magnitude of an activation and the orientation of the filter generating it. In this case,
BN should only normalize magnitudes of the vectors to unit standard deviation. It would not make sense to normalize the angles, since their values are already bounded and changing their distribution would alter important information about relative and global orientations. Given a vector field feature map $\mathbf{z}$ and its map of magnitudes $\bm{\rho}$, we compute batch normalization as:
\begin{equation}
\mathbf{\hat{z}}=\frac{\mathbf{z}}{\sqrt{\text{var}(\bm{\rho})}}.
\end{equation}

\subsection{Computational considerations}
Although RotEqNet allows for smaller models, they  might require a higher count of convolutions than a comparable standard CNN. For instance, with the architecture used for MNIST-rot in Sec.~\ref{sec:exp}, a standard CNN requires $4\times$ more filters per layer to saturate performance, compared to RotEqNet. At the same time, RotEqNet requires $R/4=4.25\times$ (for $R=17$) more convolutions. This results in RotEqNet saving $10\times$ in model memory, $2\times$ in data memory at a price of requiring just $1.5\times$ more computing time.
This is because, although the convolution count is higher, the number of feature maps per convolution is smaller. Less feature maps mean smaller convolution filters and  the possibility to use larger mini batches, both factors contributing to a faster training.

\section{Experiments}\label{sec:exp}
	
We explore the performance of RotEqNet on datasets where the orientation of the patterns of interest is arbitrary. This is very often the case in biomedical and abovehead imaging, since the orientation of the camera is usually not correlated with the patterns of interest. We apply RotEqNet to problems from these two fields, as well to MNIST-rot, a randomly rotated handwritten digit recognition benchmark. We also perform a study on the trade-off between invariance and accuracy in a synthetic patch matching problem. These case studies allow us to analyze the performance of RotEqNet in problems requiring equivariance, covariance and invariance to rotations and to analyze the effectiveness of RotEqNet to perform accurately with very small model architectures and limited training samples. 
%Based on a sensitivity analysis (see Fig.~\ref{fig:car_R}), all the results are reported using a number of orientations $R=17$ for the RotConv layers.

\begin{table}[!t]
  \centering
%  \vspace*{-3mm}
{\footnotesize 
\begin{tabular}{|c|c|}
\hline \textbf{Type} & \textbf{Size}\\ 
\rowcolor{lightgray}
\hline Input & $28 \times 28$\\
\hline RotConv, & $9\times 9$, 6 filt.\\ \cline{2-2}
$2\times2$ SP &  $14 \times 14 \times 6$ \cellcolor{lightgray} \\ 
\hline RotConv,  & $9\times 9\times 6$, 16 filt. \\  \cline{2-2}
$2\times2$ SP & \cellcolor{lightgray} $7 \times 7 \times 16$\\
\hline RotConv & $9\times 9\times 16$, 32 filt.\\ \cline{2-2}
$2\times2$ SP  & $1 \times 1 \times 32$ \cellcolor{lightgray} \\ 
\hline Fully & $1\times 1\times 32$, 128 filt.\\
connected & $1 \times 1 \times 128$ \cellcolor{lightgray}\\
\hline FC, Softmax & $1\times 1\times 128$, 10 filt.\\
\rowcolor{lightgray}
\hline Output & $1 \times 1\times 10$ \\ \hline 
\end{tabular} 
}
\vspace*{1mm}
\caption{Network architecture used on the MNIST-rot dataset. Layer parameters are in white and variables are shaded in gray.} \label{tab:mnist_arch}
\vspace*{-5mm}
\end{table}
   
\subsection{Invariance: MNIST-rot}

\begin{table}[!b]
  \vspace*{-3mm}
	\centering
	{\footnotesize
		\begin{tabular}{|c|c|}
			\hline \textbf{Method} & \textbf{Error rate (in \%)}\\ 
			\hline 
			SVM \cite{larochelle2007empirical} & 10.38$\pm$0.27\\
			%\hline 
			TIRBM~\cite{sohn2012learning} & 4.2\\
			%\hline 
			H-Net~\cite{worrall2016harmonic} & 1.69\\
			%\hline 
			ORN~\cite{zhou2017oriented} & 1.54\\
			%\hline 
			TI-pooling \cite{laptev2016ti} & 1.2\\ 
			%\hline 
			RotEqNet (Ours) & \textbf{1.09}\\ 
			\hline 
			RotEqNet, only scalar field & 2.01\\ 
			%\hline 
			RotEqNet, test-time augmentation & \textbf{1.01}\\ 
			\hline 
		\end{tabular} 
	}
	\vspace{0.1cm}
	\caption{Error rate on the MNIST-rot dataset trained on the train-val subset. }
	%\vspace*{-2mm}
	\label{tab:mnist}
\end{table}

MNIST-rot~\cite{larochelle2007empirical} is a variant of the original MNIST digit recognition dataset, where a random rotation between $0^\circ$ and $360^\circ$ is applied to each $28\times 28$ digit image. The training set is also considerably smaller than the standard MNIST, with 12k samples, from which 10k are used for training and 2k for validation. The test set consists of 50k samples. Since we aim at predicting the correct label independently from the rotation, this problem requires rotation invariance.
	
\paragraph{Model:} We test four CNN models with the same architecture, but different number of filters per layer. The largest model we used is shown in Table~\ref{tab:mnist_arch} and involves 100k parameters. The models are trained for 90 epochs, starting with a learning rate of 0.1 and reducing it gradually to 0.001. The weight decay is kept constant at 0.01. We use a dropout rate of 0.7 in the fully connected layer and batch normalization before every convolutional layer. The number of orientations is set to $R=17$.
\vspace*{-3mm}
\paragraph{Test time data augmentation:} We observe an important contribution of data augmentation at test time, a technique often used with approximately invariant or equivariant CNNs~\cite{fakhry2016deep,hu2015face}. In particular, we input to the network several rotated versions of the same image using fixed angles between $0^\circ$ and $90^\circ$. Rotation-based data augmentation at test time might seem counter-intuitive in a rotation invariant model, but the different rotations coupled to resampling of images and filters (cf. Sec.~\ref{met:rot_conv}) will produce slightly different activations. The final prediction is given by the average of such scores. We report results obtained with and without this type of augmentation.
\vspace*{-3mm}

\paragraph{Comparison to data augmented training:} In order to disentangle the contributions of data augmentation and RotEqNet, we trained the RotEqNet model and a standard CNN with the same architecture and $10\times$ more parameters. In Tab.~\ref{tab:mnist_aug}, we show the results for these models trained on both MNIST-rot and 10k digits from the original MNIST, with and without data augmentation. We observe how both methods complement each other.

\begin{figure}[!t]
%\vspace*{-2mm}
	\centering
	\includegraphics[width=0.95\linewidth]{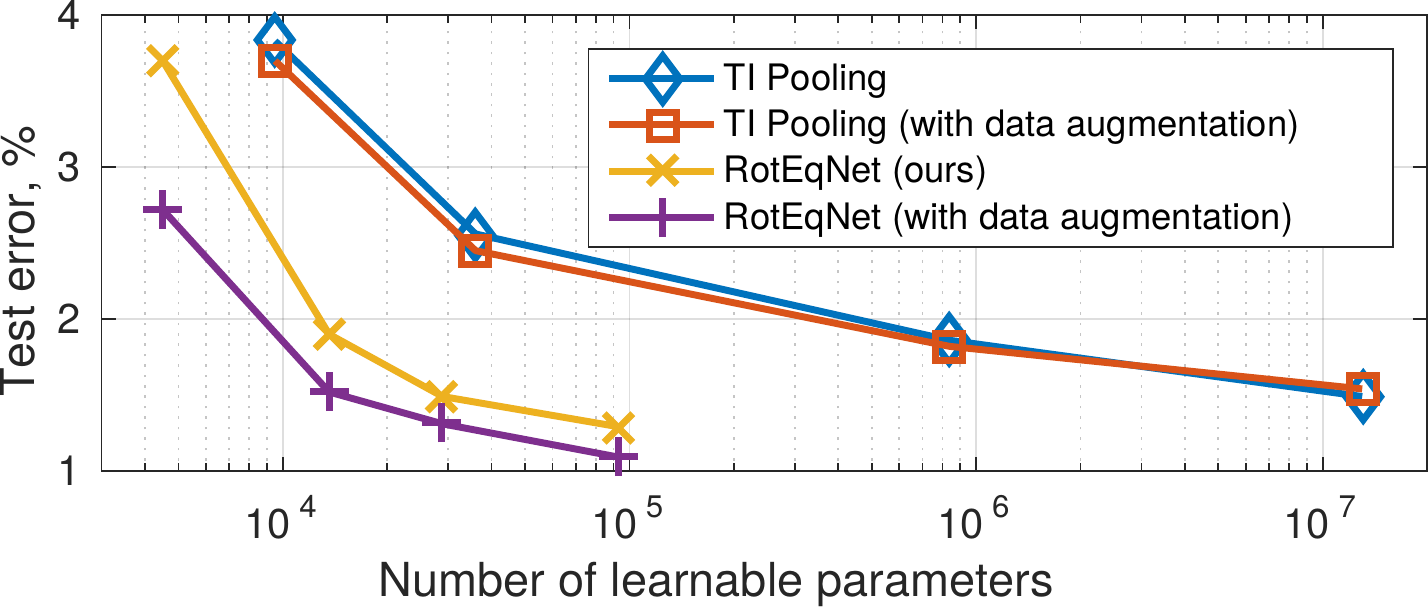}
    \caption{Performance of RotEqNet and TI-Pooling on MNIST-rot with respect to the number of parameters.}
    \label{fig:params}
    \vspace{-3mm}
\end{figure}

\begin{table}[h]
\label{tab:mnist_aug}
\begin{tabular}{c|c|c|c|c|}
\cline{2-5}
& \multicolumn{2}{c}{Train on MNIST} & \multicolumn{2}{|c|}{Train on MNIST-rot}\\
\cline{2-5}
& No augm. & Augm. & No augm. & Augm.\\
\cline{1-5}
\multicolumn{1}{|c|}{CNN} & $57\%$ & $2.3\%$ & $4.9\%$ & $2.2\%$\\
\cline{1-5}
\multicolumn{1}{|c|}{RotEqNet} & $20\%$ & $1.1\%$ & $1.4\%$ & $1.1\%$ \\
\cline{1-5}
\end{tabular}
\vspace{0.01cm}
\caption{Results on MNIST and MNIST-rot using a standard CNN or RotEqNet, with and without data augmentation.}
\vspace*{-2mm}
\end{table}
\paragraph{Results:} We first studied the behavior of RotEqNet with respect to the total number of parameters and compared it to the state-of-the-art TI-pooling~\cite{laptev2016ti}. Figure~\ref{fig:params} shows the results for both methods trained on the training set with different model sizes. The latter was achieved by varying the number of filters per layer, keeping the same architecture. RotEqNet requires approximately two orders of magnitude less parameters to obtain the same accuracy as TI-Pooling.

We report the test error in Table~\ref{tab:mnist}. RotEqNet obtains an error of $1.09\%$, a small improvement with respect to the state-of-the-art TI-pooling~\cite{laptev2016ti}, but with almost 100$\times$ less parameters. Test-time data augmentation further reduces the error to $1.01\%$, thus improving significantly over TI-Pooling and over the more recent H-Net~\cite{worrall2016harmonic} and ORN~\cite{zhou2017oriented}.

\subsection{Equivariance: ISBI 2012 Challenge}

\begin{table}[b]
\vspace*{-3mm}
  \centering
{
\footnotesize
\begin{tabular}{|c|c|}
	\hline \textbf{Type} & \textbf{Size}\\ 
	\rowcolor{lightgray}
	\hline Input & $512 \times 512$\\
	\hline RotConv, & $9\times 9$, $N$ filt.\\ \cline{2-2}
    OP, $2\times2$ SP     &  $256 \times 256 \times N \times 2$ \cellcolor{lightgray} \\ 
	\hline RotConv, & $9\times 9$, $2N$ filt.\\ \cline{2-2}
	OP, $2\times2$ SP     &  $128 \times 128 \times $2N$ \times 2$ \cellcolor{lightgray} \\ 
	\hline RotConv & $9\times 9\times $2N$ \times 2$, $3N$ filt.\\ \cline{2-2}
	OP, $2\times2$ SP  & $64 \times 64 \times $3N$ \times 2$ \cellcolor{lightgray} \\ 
	\hline RotConv, OP  & $9\times 9\times $3N$ \times 2$, $4N$ filt. \\
	Upsample and stack & $512 \times 512 \times 10N$ \cellcolor{lightgray}\\
    \hline RotConv fully & $1\times 1\times 10N \times 2$, $5N$ filt.\\
	connected & $512 \times 512 \times 5N$ \cellcolor{lightgray}\\
    \hline RotConv  & $9\times 9\times $5N$ \times 2$, $4N$ filt. \\
	OP & $512 \times 512 \times 4N$ \cellcolor{lightgray}\\
    \hline Fully   & $1\times 1\times $4N$ \times 2$, $8N$ filt. \\
	connected & $512 \times 512 \times 8N$ \cellcolor{lightgray}\\
	\hline FC, Normalize & $1\times 1\times 8N$, 3 filt.\\
	\rowcolor{lightgray}
	\hline Output & $512 \times 512\times 3$ \\
	\hline 
\end{tabular} 
}
\vspace{0.15cm}
\caption{Network architecture used with ISBI 2012 challenge data. Layer parameters are in white and variables are shaded in gray.}
\label{tab:isbi_arch}
\vspace*{-4mm}
\end{table}

This benchmark~\cite{arganda2015crowdsourcing} involves segmentation of neuronal structures in electron microscope (EM) stacks~\cite{cardona2010integrated}. In this problem we need to precisely locate the neuron membrane boundaries. Therefore, a rotation of the inputs should lead to the same rotation in the output, making the ISBI 2012 problem a good candidate to study rotation equivariance.

\begin{figure}[h]
\begin{tabular}{cccc}
\includegraphics[width=0.20\linewidth,trim={0 0 382px 312px},clip]{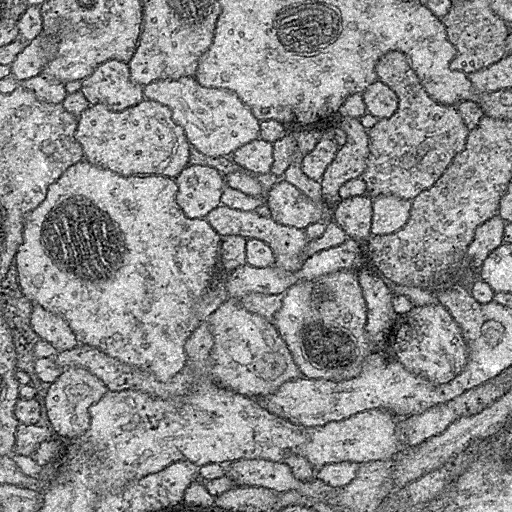} & \includegraphics[width=0.20\linewidth,trim={0 0 382px 312px},clip]{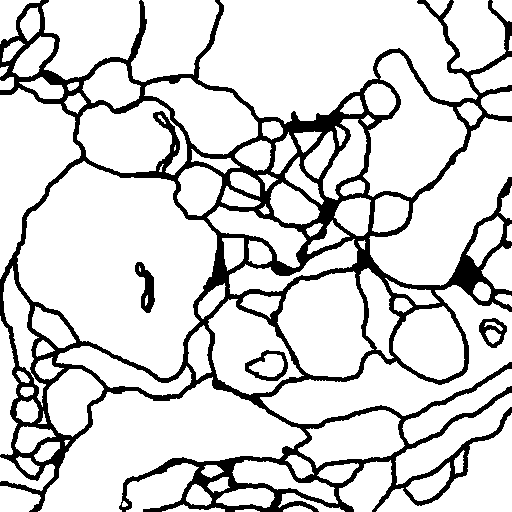} &
\includegraphics[width=0.20\linewidth,trim={0 0 382px 312px},clip]{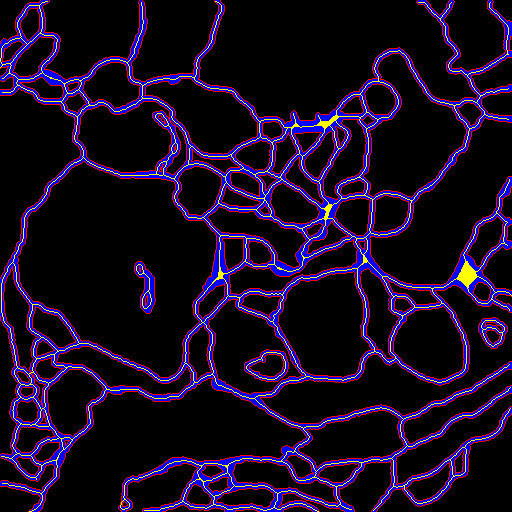} &
\includegraphics[width=0.20\linewidth,trim={0 0 382px 312px},clip]{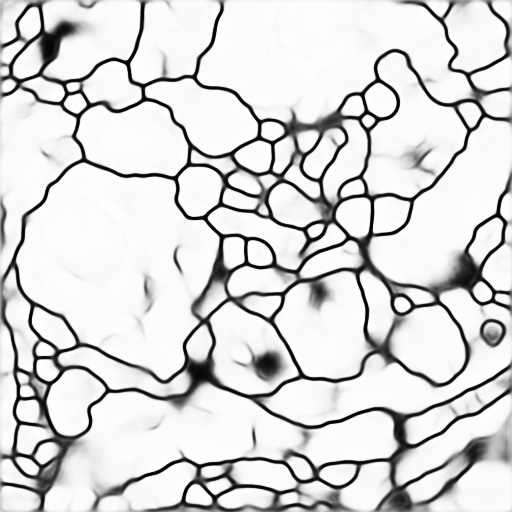}\\
(a) & (b) & (c) & (d)\\	%\multicolumn{2}{c}{\includegraphics[width=0.7\linewidth]{figs/gt30bis.png}}\\
\end{tabular}
\caption{Example validation image (\#30) of the ISBI 2012 challenge. (a) Image ($190\times130$ pixels). (b) Membrane ground truth. (c) The pre-processed 3-class ground truth: black is non-membrane, yellow is membrane center, red is membrane border and blue is non-class. (d) Probability map produced by RotEqNet.}
\label{fig:isbi}
\vspace*{-4mm}
\end{figure}

The data consist of two EM stacks of \textit{drosophila} neurons, each composed of 30 images of size $512\times 512$ pixels (Fig.~\ref{fig:isbi}a). One stack is used for training and the other for testing. The ground truth for the training stack consists of densely annotated binary images (Fig.~\ref{fig:isbi}b). The ground truth for the test stack is private and the results are to be submitted to an evaluation server~\footnote{\label{web:isbi}http://brainiac2.mit.edu/isbi\_challenge/}.

%\subsubsection{Pre-processing of the labels}

\paragraph{Model:} We transform the original binary 
problem into a three class segmentation problem: 1) non-membrane, 2) central membrane pixels and 3) external membrane pixels. Pixels in the membrane but not belonging to either 2) or 3) are considered to be unlabeled (Fig.~\ref{fig:isbi}c). This way, we can assign a higher penalization to the non-membrane pixels next to the membrane and a lower one to those in the middle of the cells. The central membrane scores are used as the final binary prediction (Fig.~\ref{fig:isbi}d).

Since we are dealing with a dense prediction problem with spatial autocorrelation at different resolution levels, we apply three RotConv blocks with spatial pooling. We then upsample the output of each block to the size of the original image, before concatenating them and applying two more RotConv blocks. Table~\ref{tab:isbi_arch} shows the  architecture. The parameter $N$ is used to change the size of the model. We evaluated the results with $N=2$ and with an ensemble of three models, with $N=[1,2,3]$.

  \paragraph{Comparison to data augmented training:} we evaluated the RotEqNet model ($N=2$) and an equivalent standard CNN with $10\times$ more parameters on 5 held out validation images. RotEqNet seems not to profit as much from data augmentation as its standard CNN counterpart, but improves the CNN solution in all the cases considered, as illustrated in Table~\ref{tab:ISBI_aug}.
\begin{table}[h!]
\centering

\label{tab:ISBI_aug}
{\footnotesize
\begin{tabular}{c|c|c|}
%& \multicolumn{2}{|c|}{Train on MNIST} & \multicolumn{2}{|c|}{Train on MNIST-rot}\\
\cline{2-3}
& No augm. & Augm. \\
\cline{1-3}
\multicolumn{1}{|c|}{CNN} & 0.9232 & 0.9572 \\
\cline{1-3}
\multicolumn{1}{|c|}{RotEqNet} & 0.9726 & 0.9790 \\
\cline{1-3}
\end{tabular}
}
\vspace*{1mm}
\caption{ISBI results on the validations set using a standard CNN or RotEqNet, with and without data augmentation.}
\end{table}

\begin{table}[!t]
\vspace*{-2mm}
  \centering
{\footnotesize
\begin{tabular}{|c|c|c|c|}
\hline \textbf{Method} & \textbf{Rand. Thin} & 	\textbf{Inf. Thin} & \# params.\\ 
\hline CUMedVision~\cite{chen2016deep} & 0.9768 & 0.9886 & -\\
IAL MC/LMC~\cite{beier2016efficient} & \textbf{0.9826} & \textbf{0.9894} & -\\
\hline 
DIVE~\cite{fakhry2016deep} & 0.9685 & 0.9858 & 5.7M\\
PolyMtl~\cite{drozdzal2016importance} & 0.9689 & 0.9861 & 11M\\			
U-Net~\cite{ronneberger2015u} & \textbf{0.9728} & \textbf{0.9866} & 33M\\
\hline RotEqNet ($N=2$) & 0.9599 & 0.9806 & 30k\\
RotEqNet, 3 models & 0.9712 & 0.9865 & 100k\\
\hline 
\end{tabular}
}
\vspace{0.1cm}
\caption{Scores on the held out test set of the ISBI 2012 Challenge. %The first two~\cite{beier2016efficient,chen2016deep} apply extended post-processing while the rest, including ours, use raw CNN probabilities.
}
%\vspace*{-3mm}
\label{tab:isbi}
\end{table}
	
\paragraph{Results:} A detailed explanation on the evaluation metrics used in the challenge can be found on the ISBI 2012 challenge website\textsuperscript{\ref{web:isbi}}, as well as in~\cite{arganda2015crowdsourcing}. The winners of the challenge were Chen \etal~\cite{chen2016deep}, although Beier \etal~\cite{beier2016efficient} have the highest scores at the time of writing.
These two works rely on complex post-processing pipeline. Our rotation equivariant prediction provides results comparable to other state-of-the-art methods only relying on the raw CNN softmax output~\cite{drozdzal2016importance,fakhry2016deep,ronneberger2015u} (see Table~\ref{tab:isbi}).

\subsection{Covariance: car orientation estimation}

Estimating car orientations from above-head imagery requires rotation covariant models. We use the dataset provided by the authors of~\cite{henriques2016warped}, which is based on Google Map images. It is composed by 15 tiles, where cars' bounding boxes and corresponding orientations come from manual annotation. We implement our approach in similarly to~\cite{henriques2016warped}. We crop a 48$\times$48 square patch around every car, based on the bounding box center point. We then use these crops for both training and testing of the model. As in~\cite{henriques2016warped}, we use the cars in the first 10 images (409 cars) for training and those in the last 5 images (209 cars) for testing. We did not use the cars whose center was nearer than 38 pixels from the image border, in order to avoid artifacts.

\begin{table}[h]
\centering
{\footnotesize
\begin{tabular}{|c|c|}
\hline \textbf{Type} & \textbf{Size}\\ \rowcolor{lightgray} \hline         
Input   & $48 \times 48$\\ \hline 
RotConv & $11\times 11$, 3 filt.\\ \cline{2-2}
OP      &  $38 \times 38 \times 4 \times 2$ \cellcolor{lightgray} \\ \hline 
RotConv & $11\times 11 \times 3 \times 2$, 6 filt.\\ \cline{2-2}
OP      &  $28 \times 28 \times 6 \times 2$ \cellcolor{lightgray} \\ \hline 
RotConv & $11\times 11\times 6 \times 2$, 3 filt.\\ \cline{2-2}
OP, $2\times2$ SP  & $9 \times 9 \times 3 \times 2$ \cellcolor{lightgray} \\ \hline
  RotConv fully & $9\times 9\times 3 \times 2$, 1 filt.\\
  connected (FC1) & $1 \times 1 \times 21$ \cellcolor{lightgray}\\ \hline
  FC2, Hardcoded & $1\times 1\times 21$, 2 filt.\\ \rowcolor{lightgray} \hline 
  Output & $1 \times 1\times 2$ \\ \hline 
\end{tabular} 
}
\vspace{0.2cm}
\caption{Architecture of the car orientation estimation network.% 9k network used for car orientation estimation.
Parameters are in white and variables are shaded in gray.}
\label{tab:cars_arch}
\vspace*{-2mm}
\end{table}

\begin{figure}[h]
\centering
\begin{tabular}{c}
\includegraphics[width=0.93\linewidth]{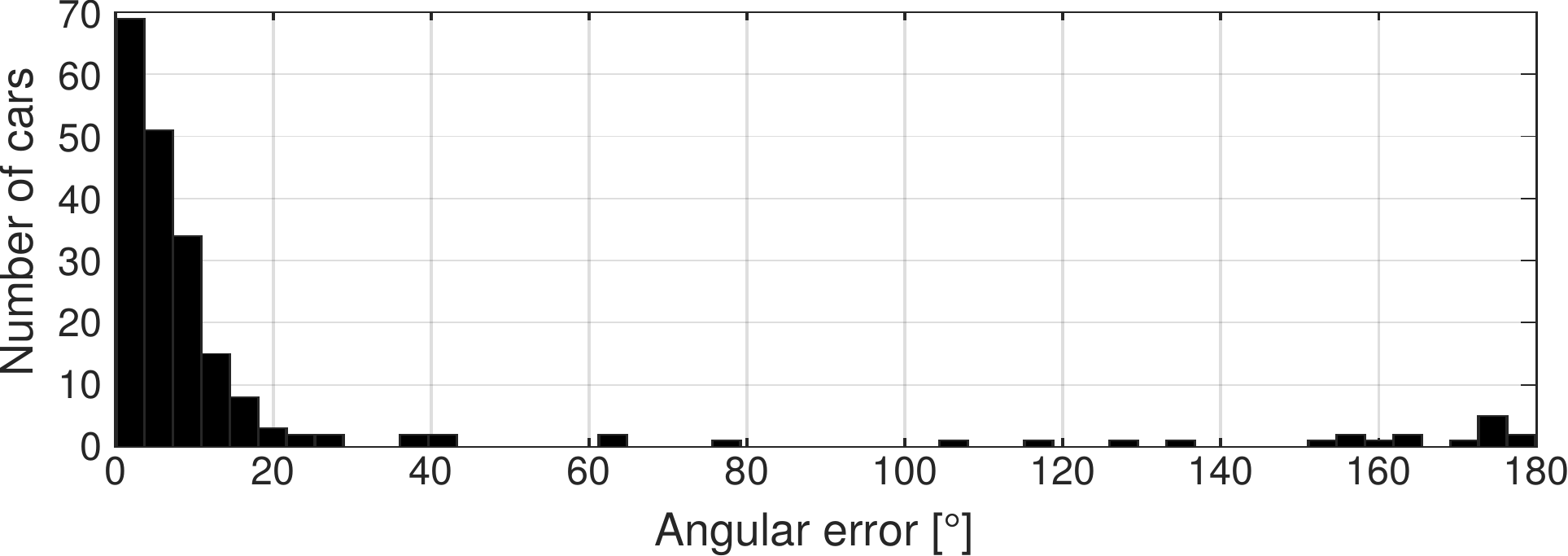} \\[-0.75mm]
\includegraphics[width=0.93\linewidth]{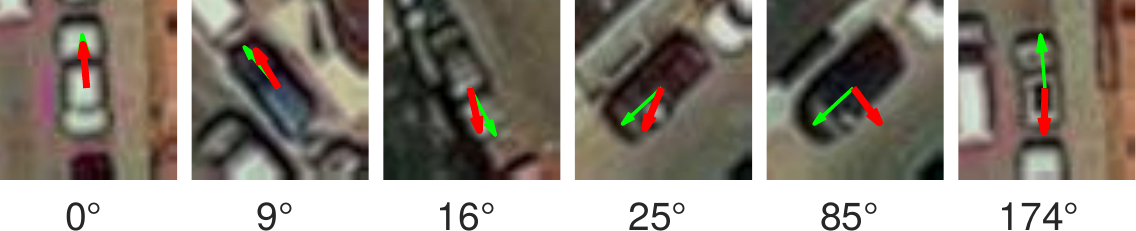}\\
\end{tabular}
\caption{Distribution of the errors in the test set (top). Examples (bottom) of correctly and incorrectly identified orientations. Ground truth arrows in green (thin) and predictions in red (thick).}
\label{fig:cars}
%\vspace{-2mm}
\end{figure}

\paragraph{Model:} We want to learn a covariant function with respect to rotations, since a rotation by $\Delta\alpha^\circ$ in the input image results in a change by $\Delta\alpha^\circ$ in the predicted angle. In particular, we train on sine and cosine of $\alpha^\circ$, since they are continuous with respect to $\Delta\alpha^\circ$.
	%\mitch{We use the dataset provided by the authors of~\cite{henriques2016warped}, which is based on images issued from Google Maps.}{} As in~\cite{henriques2016warped}, we use the cars in the first 10 images to train the network and those in the last 5 to test. 
The network's architecture is illustrated in Table~\ref{tab:cars_arch}. For the output we use a tanh non-linearity, followed by a normalization of the output vector to unit-norm. The first fully connected layer (FC1) is a RotConv block with a single filter ($R=21$) \emph{not} followed by by an Orientation Pooling, meaning that the subsequent feature vector has 21 dimensions instead of just one. We can expect this vector to undergo a circular shift when the input image is subject to a rotation. We hardcode the two mappings of the following layer (FC2) to $[\sin(360/R),\sin(2\cdot 360/R),\dots \sin(R\cdot 360/R)]$ and $[\cos(360/R),\cos(2\cdot 360/R),\dots \cos(R\cdot 360/R)]$.
This ensures that there will be no preferred orientations inherited from a biased training set. The weight decay and learning rate are $10^{-2}$ and $5\cdot 10^{-3}$ respectively, for the 80 epochs. All the filters were initialized from a normal distribution with zero mean and $\sigma=10^{-3}$. The final models correspond to the average of the weights of the last 30 epochs.

\paragraph{Results:}  Table~\ref{tab:car_res} reports the average test error. The use of RotEqNet substantially improves the results, outperforming by more than 20\% the previous state-of-the-art method~\cite{henriques2016warped}. In Fig.~\ref{fig:cars}, we show the error distribution in the test set for the hybrid model. Note how most samples, $82.7\%$, are predicted with less than $15^\circ$ of orientation error, while most of the contribution to the total error comes from the $6.7\%$ of samples with errors larger than $150^\circ$, in which the front of the car has been mistaken with the rear.

\begin{table}[h]
%\vspace*{-3mm}
  \centering
{\footnotesize
\begin{tabular}{|c|c|c|}
\hline 
\textbf{Method} & \textbf{Avg. error ($^\circ$)} & \# params\\ \hline CNN~\cite{henriques2016warped} & 28.87 & 27k\\ 
Warped-CNN~\cite{henriques2016warped} & 26.44 & 27k\\ \hline 
RotEqNet (Ours) & 24.07 & 5k \\ 
RotEqNet (Ours) & \textbf{20.46} & 9k \\ \hline 
\end{tabular} 
}
\vspace{0.2cm}
\caption{Mean error in the prediction of car orientations.}
\label{tab:car_res}
\vspace*{-3mm}
\end{table}
%
%\begin{table}[!t]
%\setlength{\tabcolsep}{1.5pt}
%\centering
%{\small
%\begin{tabular}{|c|c|c|c|c|c|c|c|c|c|c|}
%	\hline 
%    {\bf {\it R}} & 12 & 13 &   14    & 15    & 16    & 17    & 18    & 19    & 20    & %21\\ \hline 
%    {\bf Error} & 23.9 & 22.7 & 22.0 & 22.0 & 22.5 & 21.4 & 21.8 & 21.4 & 23.1 & 21.7\\ %\hline 
%\end{tabular} 
%}
%\vspace{2mm}
%\caption{Sensitivity to the number of angles used by RotConv, $R$}
%\label{tab:car_R}
%\vspace*{-3mm}
%\end{table}
	
\begin{figure}[h]
\centering
\includegraphics[width=0.85\linewidth]{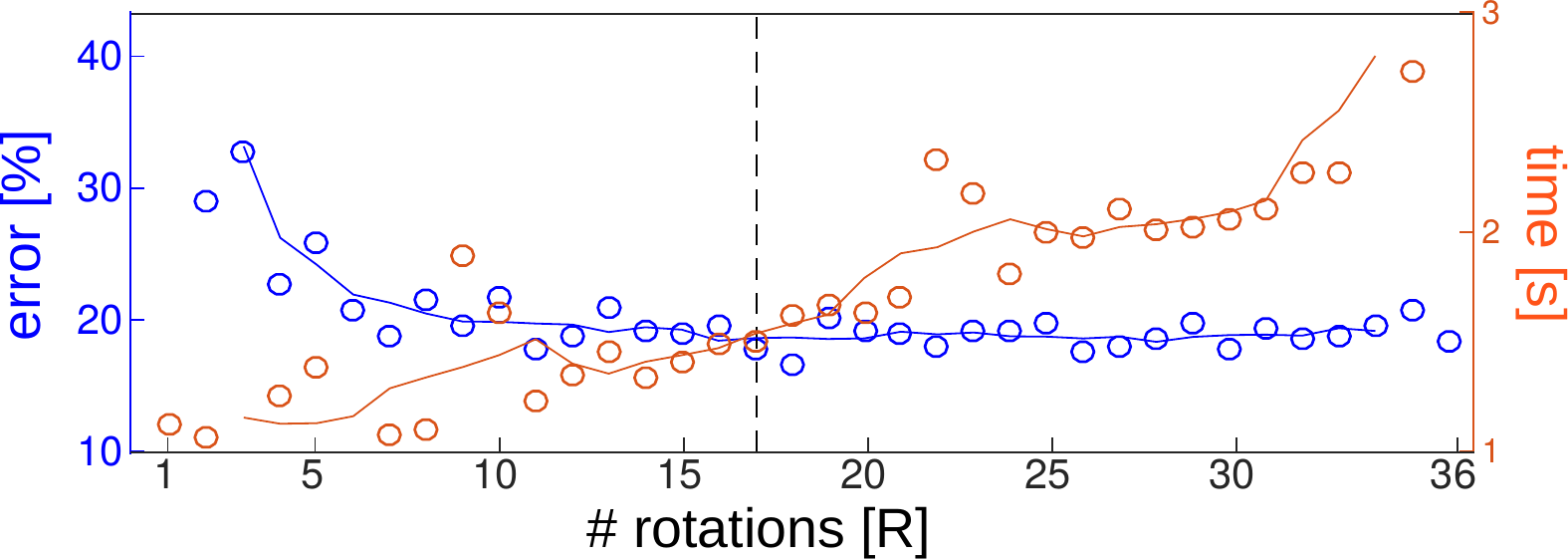} \\[-0.75mm]
\caption{Error (left y-axis, blue) vs computational time (right y-axis, red) for the number of filters considered. The vertical dashed line denotes $R=17$.}
\label{fig:car_R}
\vspace{-2mm}
\end{figure}

\vspace*{-2mm}
\paragraph{Sensitivity to $R$:} In order to study the sensitivity of RotEqNet to the number of angles $R$, we trained the model using $R=21$ and tested it for different values (see Figure~\ref{fig:car_R}). We observed relatively small changes in the test error for $R>17$. 

\subsection{Invariance 2: robustness in patch matching}

Patch matching is widely used in many image processing and computer vision problems, such as registration, 3D reconstruction and inpainting. The aim is to find matching pairs of patches (\emph{e.g.} the same features in the two different images of the same object). In this setting, the differences in orientation are often considered to be a nuisance. Although handcrafted features such as SIFT are still widely used as baselines to measure similarity, recent works have shown that learning ad-hoc features with siamese CNNs~\cite{simo2015discriminative,zagoruyko2015learning} can perform substantially better.%, when trained on a large dataset.
In the following, we apply RotEqNet to analyze how this problem can benefit from a tunable amount of rotation invariance.

Depending on the problem at hand, one might have a prior on how much rotation invariance is required. Although CNN-based descriptors are more robust to relative rotations between matching pairs than SIFT, they still tend to perform poorly for large angular differences~\cite{simo2015discriminative}.

To showcase how RotEqNet allows to tune the amount of rotation invariance, we trained a siamese network with three RotConv blocks, with 3, 6 and 32 filters of size $9\times 9$ respectively, totaling 40k parameters. The last fully connected block provides 32 scalar features. We trained it on 20k samples from the Notredame dataset~\cite{winder2007learning} with a distance-based objective function~\cite{simo2015discriminative,zagoruyko2015learning}.

\begin{figure}[h]
\centering
\begin{tabular}{c}
\includegraphics[width=0.99\linewidth]{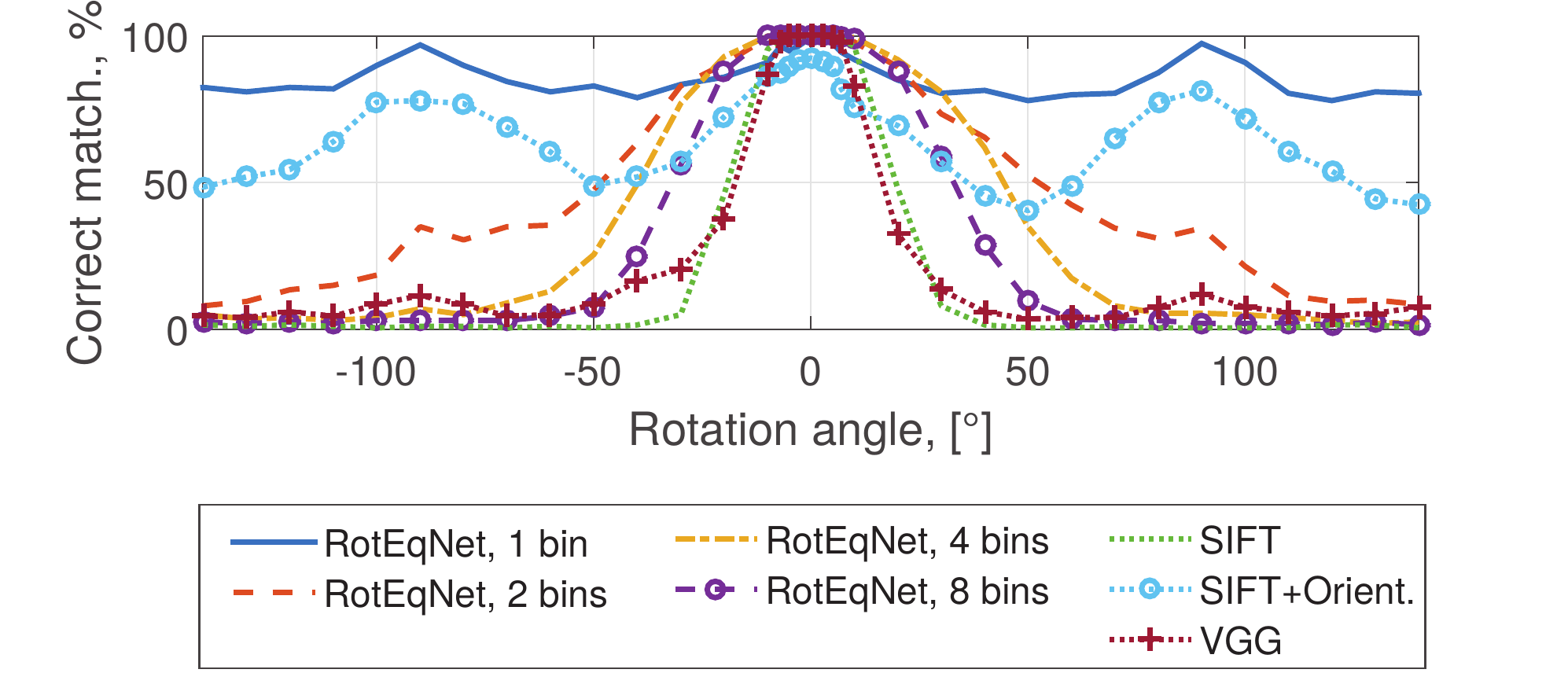} \\
\end{tabular}
\caption{Matching accuracy vs. rotation applied to one of the elements in each matching pair in a synthetic dataset. RotEqNet allows to trade-off some accuracy at small rotations for more robustness by changing the number of bins in the last Orientation Pooling layer.}
\label{fig:matching}
\vspace*{-2mm}
\end{figure}

After training, the number of bins in the last Orientation Pooling layer can be modified, thus yielding multiple descriptors per sample. For instance, if the number of bins is set to 4, one 32-dimensional descriptor will be produced for each quadrant, thus resulting in a 128-dimensional descriptor for the patch. We analyze robustness in patch matching by increasing the rotation of the patches and the number of bins, and compare our results to those obtained by SIFT and the features from a pre-trained VGG network~\cite{simo2015discriminative}. We use patches extracted from an urban photograph that are then paired to a shifted (by one pixel) and rotated version of itself. Results in Fig.\ref{fig:matching} show that RotEqNet with a single bin is much more robust to rotations than VGG and SIFT descriptors, even when the main orientation assignment is used. As a trade-off, it performs slightly worse for small rotations. However, by increasing the number of bins we can invert this tendency and improve the matching accuracy for small angles (and trade off accuracy on large rotations): using two bins (i.e. a 64-dimensional descriptor), we clearly outperform the baselines on small angles and still have 60\% of correct matches for rotations around $45^\circ$ (compared to less than $10\%$ for SIFT and VGG).

\section{Limitations and future work}
Forcing the Orientation Pooling block to choose the most activating orientation could result in exacerbating noise when there is no main orientation on either the input or the filter. This is because the arbitrarily chosen orientation can have a big impact on the output, and how it will interact with filters in the following layer, but no meaning. This problem is amplified by the use of scalar products between the vector elements of the filter and its input, which assumes that the orientation of these vectors is relevant. This issue could be improved by using a custom similarity metric between vector elements such that symmetries in the filters or the input are taken into account.

\section{Conclusion}
	
We have presented a new way of hard-coding into CNNs predefined behaviors with respect to rotations. This is achieved by applying each filter at different orientations and extracting a vector field feature map, encoding the maximum activation in terms of magnitude and angle. 

Experiments on classification, segmentation, orientation estimation and matching show the suitability of this approach for solving a wide variety of problems that are inherently rotation equivariant, invariant or covariant. These results suggest that taking into account only the dominant orientations is sufficient to tackle successfully a range of problems.
	
{\small
	\bibliographystyle{ieee}
	\bibliography{bib}
}
	
\end{document}